\documentclass{article}

\usepackage{arxiv}

\usepackage[utf8]{inputenc} 
\usepackage[T1]{fontenc}    
\usepackage{hyperref}       
\usepackage{url}            
\usepackage{booktabs}       
\usepackage{amsfonts}       
\usepackage{nicefrac}       
\usepackage{microtype}      
\usepackage{lipsum}		
\usepackage{graphicx}
\usepackage{float}
\usepackage{amsmath}
\usepackage[numbers]{natbib}
\usepackage{doi}
\usepackage{algorithm}
\usepackage{algpseudocode}

\title{Three Methods, One Problem: Classical and AI Approaches to No-Three-in-Line}

\author{ \href{https://orcid.org/0009-0000-0270-7112}{\includegraphics[scale=0.06]{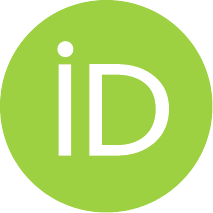}\hspace{1mm}Pranav Ramanathan} \\
	School of Mathematical Sciences\\
	Queen Mary University of London\\
	\texttt{p.ramanathan@se23.qmul.ac.uk}
	\And
	\href{https://orcid.org/0000-0003-0519-6762}{\includegraphics[scale=0.06]{orcid.pdf}\hspace{1mm}Thomas Prellberg} \\
	School of Mathematical Sciences\\
	Queen Mary University of London\\
	\texttt{t.prellberg@qmul.ac.uk}
	\And
	{\hspace{1mm}Matthew Lewis} \\
	School of Mathematical Sciences\\
	Queen Mary University of London \\
	\texttt{matthew.lewis@qmul.ac.uk}
	\And
	\href{https://orcid.org/0009-0008-5476-9962}{\includegraphics[scale=0.06]{orcid.pdf}\hspace{1mm}Prathamesh Dinesh Joshi} \\
	Vizuara AI Labs \\
	\texttt{prathamesh@vizuara.com}
	\And
	{\hspace{1mm}Raj Abhijit Dandekar} \\
	Vizuara AI Labs \\
	\texttt{raj@vizuara.com}
	\And
	{\hspace{1mm}Rajat Dandekar} \\
	Vizuara AI Labs \\
	\texttt{rajatdandekar@vizuara.com}
	\And
	{\hspace{1mm}Sreedath Panat} \\
	Vizuara AI Labs \\
	\texttt{sreedath@vizuara.com}
}

\hypersetup{
pdftitle={Three Methods, One Problem: Classical and AI Approaches to No-Three-in-Line},
pdfsubject={cs.AI,math.CO},
pdfauthor={Pranav Ramanathan},
pdfkeywords={No-Three-In-Line, combinatorial optimization, integer linear programming, transformer models, reinforcement learning, PatternBoost, constraint satisfaction},
}

\begin{document}
\maketitle

\begin{abstract}
The No-Three-In-Line problem asks for the maximum number of points that can be placed on an $n \times n$ grid with no three collinear, representing a famous problem in combinatorial geometry. While classical methods like Integer Linear Programming (ILP) guarantee optimal solutions, they face exponential scaling with grid size, and recent advances in machine learning offer promising alternatives for pattern-based approximation. This paper presents a systematic comparison of classical optimization and AI approaches to this problem, evaluating their performance against traditional algorithms. We apply PatternBoost transformer learning and reinforcement learning (PPO) to this problem for the first time, comparing them against ILP. ILP achieves provably optimal solutions up to $19 \times 19$ grids, while PatternBoost matches optimal performance up to $14 \times 14$ grids with 96\% test loss reduction. PPO achieves perfect solutions on $10 \times 10$ grids but fails at $11 \times 11$ grids, where constraint violations prevent valid configurations. These results demonstrate that classical optimization remains essential for exact solutions while AI methods offer competitive performance on smaller instances, with hybrid approaches presenting the most promising direction for scaling to larger problem sizes.
\end{abstract}

\keywords{combinatorial optimization \and constraint satisfaction \and integer linear programming \and No-Three-In-Line \and PatternBoost \and reinforcement learning \and transformer models}

\section{Introduction}

\begin{figure}[htbp]
\centering
\includegraphics[width=0.55\textwidth]{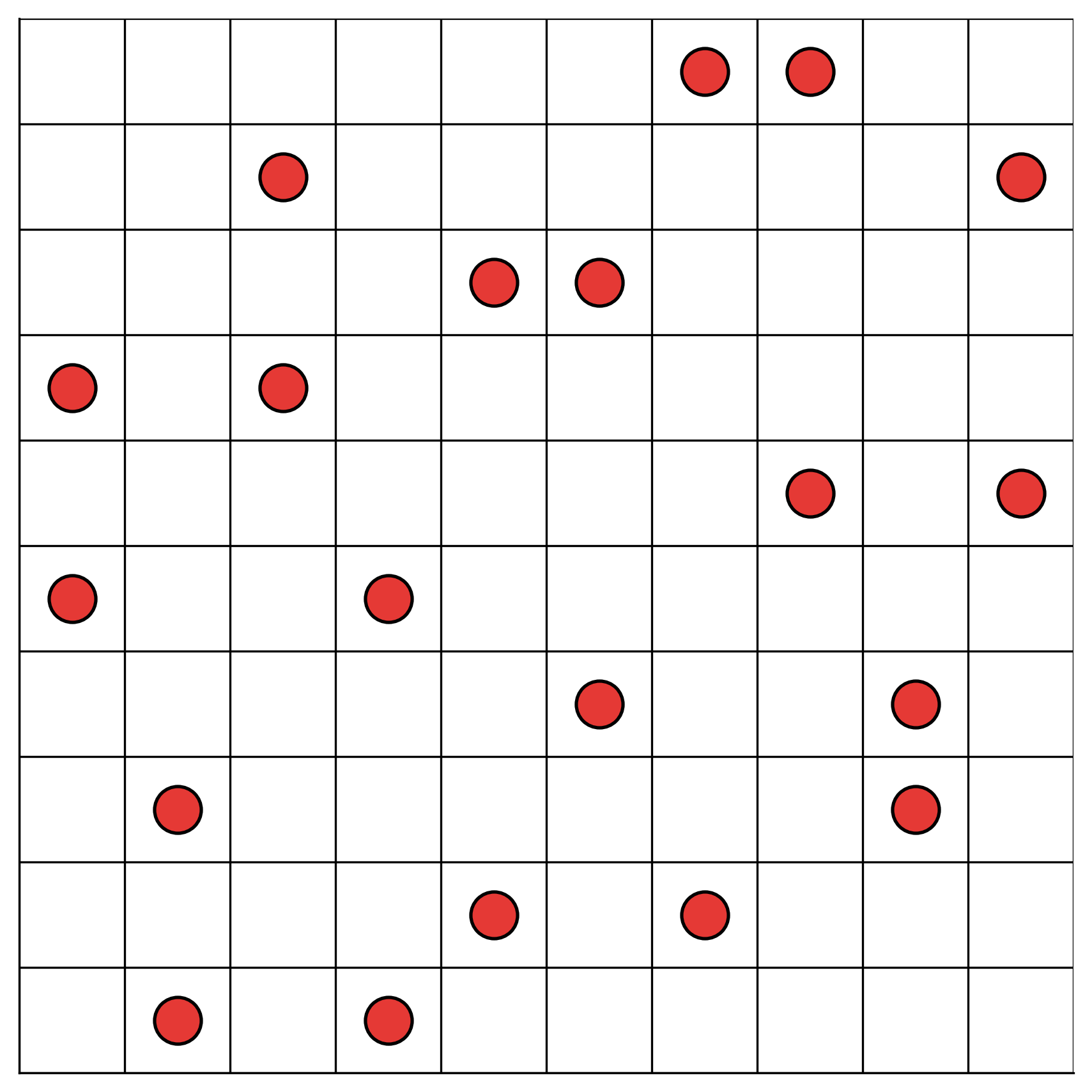}
\caption{Optimal solution for the No-Three-In-Line problem on a $10 \times 10$ grid, achieving the maximum of 20 points.}
\label{fig:problem_intro}
\end{figure}

We consider the No-Three-In-Line problem \cite{GuyKelly1968}: What is the maximum number of points that can be placed on an $n \times n$ grid such that no three are collinear? Figure \ref{fig:problem_intro} illustrates an optimal solution for a $10 \times 10$ grid, achieving the maximum of 20 points with no three collinear.

A trivial upper bound of $2n$ points follows from the pigeonhole principle, as no row or column can contain more than two points. The first non-trivial lower bound was established by Paul Erd{\H{o}}s, who used modular parabola constructions to place approximately $n$ points \cite{ErdosRoth1951}. This was improved to approximately $1.5n$ by Hall, Jackson, Sudbery, and Wild using modular hyperbola constructions, which remains the best proven lower bound \cite{Hall1975}. Exact solutions achieving $2n$ points have been found for all $n \leq 58$\footnote{T.~Prellberg, private communication, 2025.}. However, Guy and Kelly conjectured that for sufficiently large $n$, the maximum asymptotically approaches only $cn$ points, where $c \approx 1.814$ \cite{GuyKelly1968}. This leaves a significant gap between the best-proven result and the conjectured limit, underscoring the problem's persistent difficulty.

Traditional computational approaches, such as recursive backtracking algorithms and integer linear programming (ILP), frame the No-Three-In-Line problem as a constraint satisfaction task by encoding collinearity constraints algebraically and using branch-and-bound optimization \cite{NemhauserWolsey1988, Eppstein2018}. Early computational efforts by Flammenkamp employed exhaustive backtracking search with sophisticated pruning techniques, successfully determining optimal configurations up to $n = 52$ \cite{Flammenkamp2010}. These enumeration methods systematically explore the search space while eliminating symmetric equivalents and configurations that violate constraints, but require exponentially increasing computation time as grid size grows. These classical methods face two fundamental scaling challenges. First, the search space grows exponentially as $2^{n^2}$, making exhaustive search futile for larger $n$. Second, the number of collinearity constraints that must be checked grows quadratically with each new point placed, making each new point placement increasingly expensive.

Recent advances in machine learning offer promising alternatives for combinatorial optimization. Reinforcement learning methods have demonstrated success across diverse combinatorial domains, including routing, scheduling, and game-playing tasks \cite{Mazyavkina2021}, while graph neural networks have shown effectiveness in learning heuristics for constraint satisfaction and discrete optimization problems \cite{Cappart2023}. Meanwhile, transformer architectures \cite{Vaswani2017} are particularly well-suited to problems with global constraints, as their attention mechanisms can model relationships across all points in a set simultaneously. A recent breakthrough came with PatternBoost \cite{Charton2024}, which discovered a counterexample to Graham and Harary's conjecture on hypercube subgraphs \cite{GrahamHarary1992}, a problem that had remained unresolved for 30 years, and also improved bounds on extremal combinatorics problems including triangle-free graphs, point configurations avoiding isosceles triangles, and the no-5-in-sphere problem. While these applications demonstrate PatternBoost's effectiveness on geometric point placement problems with distance-based constraints, its application to collinearity-based constraints remains unexplored. Early work by Tsuchiya and Takefuji applied classical neural networks to the No-Three-In-Line problem in 1995 \cite{Tsuchiya1995}, but no subsequent studies have explored transformer-based pattern learning or reinforcement learning approaches, leaving a significant gap in understanding how contemporary AI methods perform on collinearity constraint satisfaction in combinatorial geometry.

This study presents the first comparison of classical and AI-driven methods for the No-Three-In-Line problem. We compare three approaches: integer linear programming using the Gurobi solver, transformer-based pattern learning via the PatternBoost framework \cite{Charton2024}, and reinforcement learning using Proximal Policy Optimization (PPO) \cite{Schulman2017}. 

\textbf{Paper structure as follows:}
\begin{itemize}
    \item Section 2 describes the implementation and training procedures for each method
    \item Section 3 presents comparative results and scaling behavior
    \item Section 4 discusses algorithmic trade-offs, failure modes, and future research directions
\end{itemize}

\section{Methodology}

\subsection{Problem Formulation}

The No-Three-In-Line problem is defined on an $n \times n$ discrete grid, where each cell $(i,j)$ with $0 \leq i,j < n$ represents a potential point placement. We represent a configuration as a binary matrix $\mathbf{X} \in \{0,1\}^{n \times n}$, where $x_{i,j} = 1$ indicates a point is placed at position $(i,j)$, and $x_{i,j} = 0$ indicates an empty cell.

\textbf{Collinearity Constraint.} Three distinct points $(x_1, y_1)$, $(x_2, y_2)$, and $(x_3, y_3)$ are collinear if and only if they satisfy:
\begin{equation}
\begin{vmatrix}
x_1 & y_1 & 1 \\
x_2 & y_2 & 1 \\
x_3 & y_3 & 1
\end{vmatrix} = 0
\end{equation}

This determinant evaluates to zero when the three points lie on a common line (horizontal, vertical, or diagonal). The constraint applies to \textit{all} possible straight lines passing through the grid, including lines with arbitrary rational slopes.

\textbf{Optimization Objective and Constraints.} With decision variables $x_{i,j} \in \{0,1\}$ indicating a point at $(i,j)$, the ILP formulation is:
\begin{align}
\max_{x_{i,j} \in \{0,1\}} \quad & \sum_{i=0}^{n-1} \sum_{j=0}^{n-1} x_{i,j} \\
\text{s.t.} \quad & \sum_{(i,j) \in \ell} x_{i,j} \le 2 \quad \forall \text{ grid lines } \ell \text{ with } |\ell| \ge 3.
\end{align}
Equivalently, the line constraint can be written for every triple of distinct grid points whose determinant above is zero:
\[
x_{i_1,j_1} + x_{i_2,j_2} + x_{i_3,j_3} \le 2 \quad \forall \text{ distinct } (i_k, j_k) \text{ with } \begin{vmatrix} i_1 & j_1 & 1 \\ i_2 & j_2 & 1 \\ i_3 & j_3 & 1 \end{vmatrix} = 0.
\]

\textbf{Constraint Enumeration.} For an $n \times n$ grid, there are $n^2$ grid cells, yielding $\binom{n^2}{3} = O(n^6)$ possible triplets of points. However, not all triplets lie on a common line. The number of \textit{collinear} triplets is $O(n^3)$, corresponding to all possible lines (defined by slope and intercept) that pass through at least three grid points. This determines the number of active constraints in the optimization problem.

This formulation establishes the combinatorial and geometric foundation upon which all subsequent methods are built.

\subsection{Integer Linear Programming}

We solve the No-Three-In-Line problem using integer linear programming (ILP) with the Gurobi Optimizer \cite{GurobiOptimization2024}. Using the formulation from Section 2.1, the implementation focuses on efficient constraint generation and solver configuration.

\textbf{Constraint Generation.} To enforce the no-three-in-line requirement, we enumerate collinear point sets as formalized in Algorithm~\ref{alg:ilp_constraints}.
\begin{algorithm}[htbp]
\caption{ILP Constraint Generation}
\label{alg:ilp_constraints}
\begin{algorithmic}[1]
\Require Grid size $n$
\Ensure Set of linear constraints $\mathcal{C}$
\State Initialize $\mathcal{C} \gets \emptyset$, line dictionary $\mathcal{L} \gets \{\}$
\For{each pair of distinct grid points $(i_1, j_1)$ and $(i_2, j_2)$}
    \State Compute line equation $ax + by + c = 0$ through both points
    \State Normalize coefficients: $(a, b, c) \gets (a, b, c) / \gcd(|a|, |b|, |c|)$
    \State Apply consistent sign convention to obtain canonical form $\ell$
    \If{$\ell \notin \mathcal{L}$}
        \State $\mathcal{L}[\ell] \gets \emptyset$
    \EndIf
    \State $\mathcal{L}[\ell] \gets \mathcal{L}[\ell] \cup \{(i_1, j_1), (i_2, j_2)\}$
\EndFor
\For{each line $\ell \in \mathcal{L}$ with $|\mathcal{L}[\ell]| \geq 3$}
    \State Add constraint: $\sum_{(i,j) \in \mathcal{L}[\ell]} x_{i,j} \leq 2$ to $\mathcal{C}$
\EndFor
\State \Return $\mathcal{C}$
\end{algorithmic}
\end{algorithm}

\textbf{Implementation.} The resulting ILP model contains $n^2$ binary variables and $O(n^3)$ linear constraints, corresponding to the number of distinct lines passing through at least three grid points. We solve the model using Gurobi's branch-and-bound algorithm \cite{NemhauserWolsey1988} with default settings for cutting planes and primal heuristics.

\subsection{PatternBoost Transformer Learning}

We apply the PatternBoost methodology \cite{Charton2024} to the No-Three-In-Line problem, using a transformer to learn patterns from high-quality configurations and guide the search toward better solutions.

\textbf{Grid Representation and Tokenization.} Each $n \times n$ grid configuration is encoded as a sequence of tokens for transformer input. We map each grid cell $(i,j)$ to a unique integer token $t_{i,j} = i \cdot n + j \in \{0, 1, \ldots, n^2-1\}$. A configuration containing points at positions $\mathcal{P} = \{(i_1, j_1), (i_2, j_2), \ldots, (i_k, j_k)\}$ is represented as the sequence of tokens $(t_{i_1,j_1}, t_{i_2,j_2}, \ldots, t_{i_k,j_k})$. Each token corresponds to a single grid cell, yielding a vocabulary of size $n^2$. We prepend a special start token to each sequence.

\begin{figure}[htbp]
	\centering
	\includegraphics[width=0.65\textwidth]{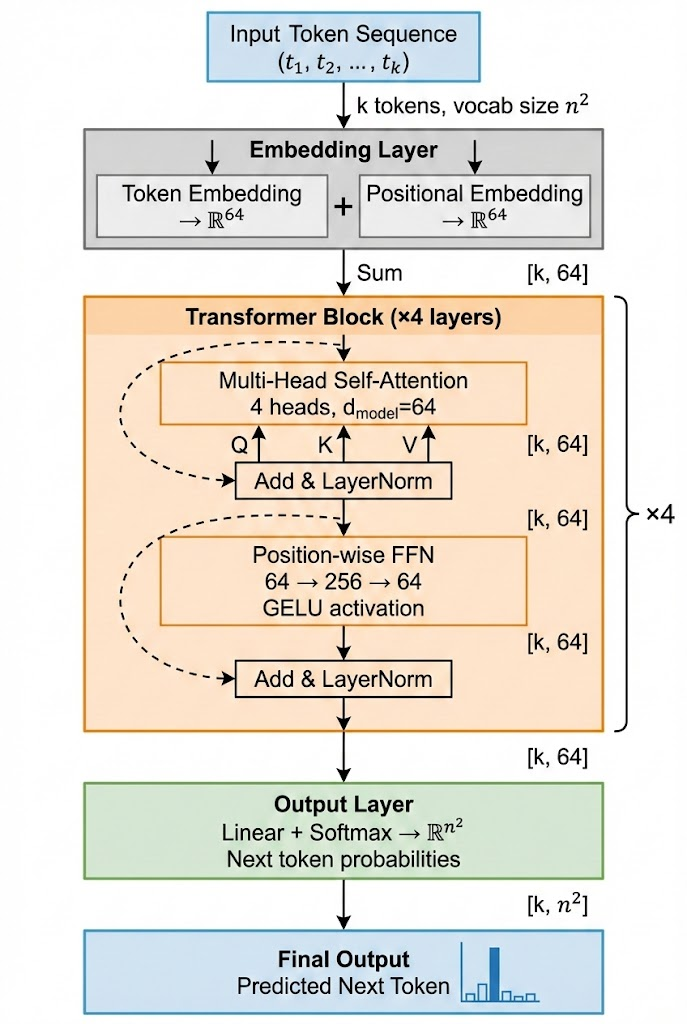}
	\caption{PatternBoost transformer: 4 layers, 4 heads, dimension 64, with residual connections.}
	\label{fig:transformer_architecture}
\end{figure}

\textbf{Transformer Architecture.} We employ a GPT-2 style decoder-only transformer \cite{Radford2019} based on Karpathy's Makemore implementation \cite{Karpathy2024}. The model consists of 4 transformer layers with 4 attention heads and a hidden dimension of 64. The feedforward network in each layer expands to dimension 256 ($4 \times$ hidden dimension) with GELU activation \cite{Hendrycks2023}. Token embeddings and learned positional embeddings are summed before processing through the transformer layers. The model performs next-token prediction via cross-entropy loss over the vocabulary. Figure \ref{fig:transformer_architecture} illustrates the complete architecture with residual connections between layers.

\textbf{Training Data Generation.} Initial training data is generated via greedy saturation from the empty grid, formalized in Algorithm~\ref{alg:greedy_saturation}.
\begin{algorithm}[htbp]
\caption{Greedy Saturation for Training Data Generation}
\label{alg:greedy_saturation}
\begin{algorithmic}[1]
\Require Empty $n \times n$ grid
\Ensure Saturated configuration with no valid moves remaining
\State Initialize grid $G \gets \emptyset$ (all cells empty)
\While{valid placements exist}
    \State $\mathcal{C} \gets \{(i,j) : G[i,j] = 0\}$ \Comment{All empty cells}
    \For{each candidate $(i,j) \in \mathcal{C}$}
        \State Simulate placing point at $(i,j)$
        \State Update forbidden regions based on collinearity constraints
        \State $\text{score}[i,j] \gets |\text{remaining available cells}|$
    \EndFor
    \State $(i^*, j^*) \gets \arg\max_{(i,j) \in \mathcal{C}} \text{score}[i,j]$ \Comment{Ties broken randomly}
    \State Place point at $(i^*, j^*)$ in $G$: $G[i^*, j^*] \gets 1$
\EndWhile
\State \Return $G$
\end{algorithmic}
\end{algorithm}
This greedy lookahead strategy produces saturated configurations. We generate 20,000 such configurations in batches of 500, forming the initial training pool.

\textbf{Priority Queue (TopPool).} Best configurations are maintained in a fixed-capacity priority queue implemented as a min-heap of size $K = 20{,}000$. Each entry stores a tuple $(s, \sigma)$ where $s$ is the configuration score (number of points) and $\sigma$ is the tokenized sequence. When the heap is full and a new configuration with score $s' > s_{\min}$ arrives, the minimum-score entry is evicted and the new entry is inserted. This ensures training always uses the top-performing configurations discovered.

\textbf{Iterative Refinement Loop.} PatternBoost alternates between local and global optimization over 20 generations, formalized in Algorithm~\ref{alg:patternboost}.
\begin{algorithm}[htbp]
\caption{PatternBoost Iterative Refinement}
\label{alg:patternboost}
\begin{algorithmic}[1]
\Require Initial pool $\mathcal{P}_0$ of 20,000 configurations, number of generations $T = 20$
\Ensure Optimized configuration pool $\mathcal{P}_T$
\For{generation $t = 1$ to $T$}
    \State \textbf{Load Pool:} Initialize TopPool with $\mathcal{P}_{t-1}$
    \State \textbf{Data Augmentation:} $\mathcal{D}_t \gets \emptyset$
    \For{each configuration $c \in \mathcal{P}_{t-1}$}
        \State Generate all 8 symmetries (4 rotations $\times$ 2 reflections)
        \State Apply canonical form deduplication
        \State Add augmented configurations to $\mathcal{D}_t$
    \EndFor
    \State \textbf{Train Transformer:} Train model $\theta_t$ for 5,000 steps on $\mathcal{D}_t$ using AdamW \cite{Loshchilov2019} with cross-entropy loss
    \State \textbf{Generate Candidates:} Sample 20,000 token sequences $\{\sigma_1, \ldots, \sigma_{20000}\}$ from trained model $\pi_{\theta_t}$
    \For{each sequence $\sigma_i$, $i = 1$ to $20000$}
        \State \textbf{Decode:} Convert token sequence $\sigma_i$ to grid coordinates
        \State \textbf{Greedy Saturation:} Apply Algorithm~\ref{alg:greedy_saturation} to complete partial configuration
        \State \textbf{Evaluate:} Compute score $s_i$ (number of points placed)
        \State Attempt to insert $(s_i, \sigma_i)$ into TopPool (min-heap with capacity 20,000)
    \EndFor
    \State \textbf{Save Generation:} Persist updated pool $\mathcal{P}_t$ to disk
\EndFor
\State \Return $\mathcal{P}_T$
\end{algorithmic}
\end{algorithm}

Table \ref{tab:patternboost_hyperparams} summarizes the complete hyperparameter configuration.

\begin{table}[htbp]
\centering
\caption{PatternBoost Hyperparameters}
\label{tab:patternboost_hyperparams}
\begin{tabular}{ll}
\toprule
\textbf{Parameter} & \textbf{Value} \\
\midrule
\multicolumn{2}{l}{\textit{Transformer Architecture}} \\
Layers & 4 \\
Attention heads & 4 \\
Hidden dimension & 64 \\
Feedforward dimension & 256 \\
Activation function & GELU \\
Loss function & Cross-entropy \\
\midrule
\multicolumn{2}{l}{\textit{Training Configuration}} \\
Optimizer & AdamW \\
Learning rate & $10^{-4}$ \\
Weight decay & 0.1 \\
Batch size & 64 \\
\bottomrule
\end{tabular}
\end{table}


\textbf{Symmetry Handling.} To increase training data diversity and prevent redundant storage, we apply symmetry processing at two stages. During data augmentation, each configuration is expanded to all 8 symmetric versions (4 rotations and 2 reflection states), increasing the training dataset size by $8\times$. When inserting into TopPool, configurations are mapped to their lexicographically smallest canonical form under these symmetries to prevent storing duplicate equivalent solutions. This approach maximizes training signal while maintaining pool diversity.


\subsection{Proximal Policy Optimization Reinforcement Learning}

We formulate the No-Three-In-Line problem as a sequential decision-making task and solve it using deep reinforcement learning. We employ Proximal Policy Optimization (PPO) \cite{Schulman2017} with action masking to train agents that learn to place points on the grid while satisfying the collinearity constraint.

\subsubsection{Markov Decision Process Formulation}

The problem is modeled as a Markov Decision Process (MDP) \cite{Towers2024} defined by the tuple $(\mathcal{S}, \mathcal{A}, P, R, \gamma)$. The state space $\mathcal{S}$ consists of $n \times n$ binary grids $s_t \in \{0,1\}^{n \times n}$, where $s_t(i,j) = 1$ indicates a placed point. The action space $\mathcal{A}$ is discrete with $|\mathcal{A}| = n^2$, where action $a = i \cdot n + j$ corresponds to placing a point at position $(i,j)$. We implement action masking to restrict the agent to empty cells only. The transition function $P(s_{t+1}|s_t, a_t)$ is deterministic, setting $s_{t+1}(i,j) = 1$ when action $a_t$ selects position $(i,j)$.

The reward function balances exploration with constraint satisfaction using sparse rewards and dense penalties:
\begin{equation}
R(s_t, a_t) = \begin{cases}
+1.0 & \text{if placement is valid (no collinearity violation)} \\
-10 & \text{if placement creates three collinear points} \\
-1 & \text{if attempting to place on occupied cell} \\
+100 & \text{if } k = 2n \text{ and violations} = 0 \text{ (perfect solution)} \\
+10 & \text{if } k = 2n \text{ and violations} > 0 \text{ (completion)}
\end{cases}
\end{equation}
where $k$ is the current number of placed points. The 10:1 penalty-to-reward ratio encourages constraint satisfaction over rapid point accumulation, while the large perfect solution bonus creates a strong gradient toward zero-violation configurations. We use discount factor $\gamma = 0.99$.

Collinearity is detected using the determinant formula: three points $(x_1, y_1)$, $(x_2, y_2)$, $(x_3, y_3)$ are collinear if $(x_2 - x_1)(y_3 - y_1) = (x_3 - x_1)(y_2 - y_1)$. For each new placement, we check all $\binom{k}{2}$ pairs of existing points. Episodes terminate when $k = 2n$ points are placed, with no early termination on violations to allow recovery from mistakes.

\subsubsection{PPO Algorithm and Network Architecture}

We employ Proximal Policy Optimization with action masking (MaskablePPO) from Stable-Baselines3 Contrib \cite{Raffin2021}. PPO optimizes a clipped surrogate objective:
\begin{equation}
L^{CLIP}(\theta) = \mathbb{E}_t \left[ \min\left(r_t(\theta)\hat{A}_t, \text{clip}(r_t(\theta), 1-\epsilon, 1+\epsilon)\hat{A}_t\right) \right]
\end{equation}
where $r_t(\theta) = \frac{\pi_\theta(a_t|s_t)}{\pi_{\theta_{\text{old}}}(a_t|s_t)}$ is the probability ratio and $\epsilon = 0.2$ is the clipping parameter. The total loss combines the clipped objective with value function loss $L^{VF}(\theta) = (V_\theta(s_t) - V^{\text{target}}_t)^2$ and entropy bonus $S[\pi_\theta](s_t)$:
\begin{equation}
L_{\text{total}}(\theta) = L^{CLIP}(\theta) - c_1 L^{VF}(\theta) + c_2 S[\pi_\theta](s_t)
\end{equation}
with coefficients $c_1 = 0.5$ and $c_2 = 0.01$. Advantage estimates $\hat{A}_t$ are computed using Generalized Advantage Estimation (GAE) \cite{Schulman2016} with $\lambda = 0.95$ and $\gamma = 0.99$.

\begin{figure}[H]
	\centering
	\includegraphics[width=0.75\textwidth]{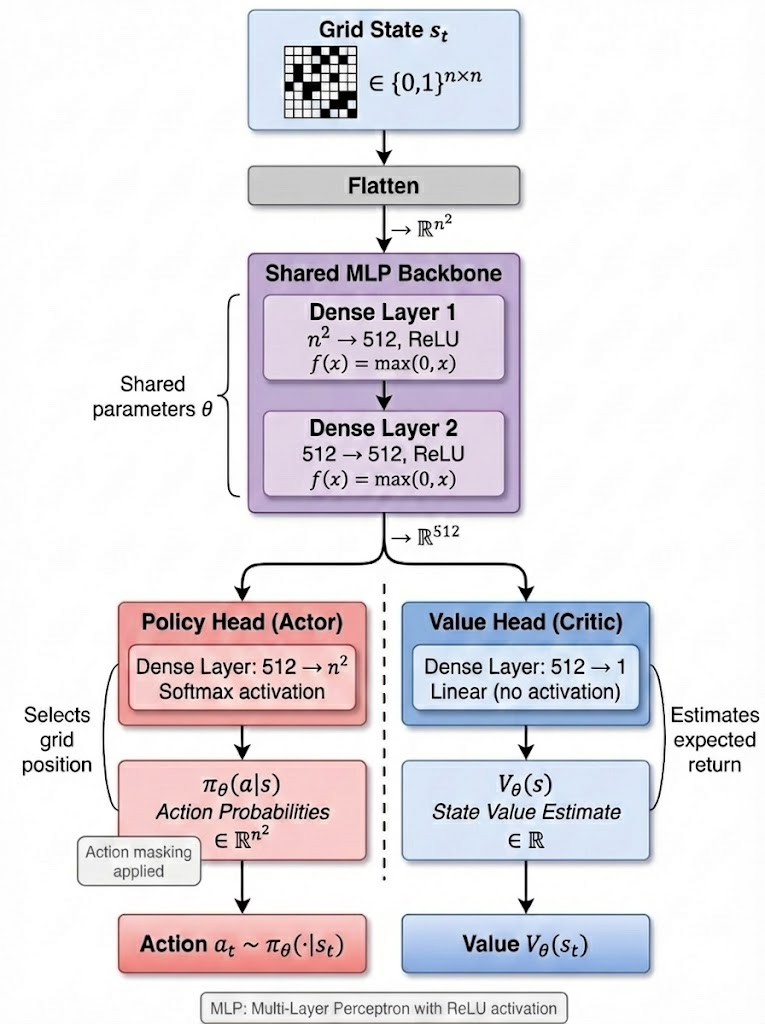}
	\caption{PPO actor-critic with shared $512\times2$ MLP backbone splitting into policy and value heads.}
	\label{fig:ppo_architecture}
	\end{figure}

The policy network uses a multi-layer perceptron (MLP) with two hidden layers of 512 units each and ReLU activations. The architecture processes the flattened $n^2$-dimensional observation vector to produce action probabilities over the $n^2$ action space. The policy network (actor) $\pi_\theta(a|s)$ and value function (critic) $V_\theta(s)$ share feature extraction layers but have separate output heads, as shown in Figure \ref{fig:ppo_architecture}.

\subsubsection{Training Configuration}

We use the Adam optimizer \cite{Kingma2015} with learning rate $\alpha_0 = 3 \times 10^{-4}$ and exponential decay: $\alpha_t = \alpha_0 \cdot (0.8)^{\lfloor t / 10^6 \rfloor}$. Training employs 30 parallel environments, collecting 2048 steps per environment (61,440 total samples) before each policy update using batches of 1024 samples. The complete training runs for 10 epochs of 1,000,000 steps each (10,000,000 total steps). Table \ref{tab:ppo_hyperparams} summarizes the complete hyperparameter configuration.

\begin{table}[htbp]
\centering
\caption{PPO Hyperparameters}
\label{tab:ppo_hyperparams}
\begin{tabular}{ll}
\toprule
\textbf{Parameter} & \textbf{Value} \\
\midrule
Training epochs & 10 \\
Network architecture & [512, 512] MLP \\
Activation function & ReLU \\
Learning rate & $3 \times 10^{-4}$ \\
Optimizer & Adam ($\beta_1=0.9$, $\beta_2=0.999$) \\
GAE $\lambda$ & 0.95 \\
Discount $\gamma$ & 0.99 \\
PPO clip $\epsilon$ & 0.2 \\
Value coefficient $c_1$ & 0.5 \\
Entropy coefficient $c_2$ & 0.01 \\
\bottomrule
\end{tabular}
\end{table}

Model performance is evaluated every epoch over 20 episodes using success rate (fraction achieving $2n$ points with zero violations), average violations, and a composite score that balances points placed, efficiency, and constraint satisfaction. The best model is selected based on maximum composite score across all training epochs.

\section{Experiments and Results}

\subsection{Experimental Setup}

We evaluated each method across multiple grid sizes to assess solution quality and scalability. ILP was tested on grids up to $n = 19$, PatternBoost on grids up to $n = 15$, and PPO on grids up to $n = 11$. All experiments were conducted on an Apple M2 with 32 GB RAM. Software implementations used Python with PyTorch for neural network training, Gurobi for integer programming, and Stable-Baselines3 for reinforcement learning.

\subsection{Solution Quality and Scalability}

Table \ref{tab:results_comparison} presents the solution quality achieved by each method across different grid sizes. We report the maximum number of points successfully placed while satisfying the no-three-in-line constraint (zero violations). For reference, we include known optimal values $2n$ for grid size $n \times n$.

\begin{table}[htbp]
\centering
\caption{Solution Quality Comparison Across Methods and Grid Sizes}
\label{tab:results_comparison}
\begin{tabular}{ccccc}
\toprule
\textbf{Grid Size} & \textbf{ILP} & \textbf{PatternBoost} & \textbf{PPO} & \textbf{Known Optimal} \\
\midrule
$5\times 5$   & 10 & 10 & 10 & 10 \\
$8\times 8$   & 16 & 16 & 16 & 16 \\
$10\times 10$ & 20 & 20 & 20 & 20 \\
$12\times 12$ & 24 & 24 & - & 24 \\
$14\times 14$ & 28 & 28 & - & 28 \\
$15\times 15$ & 30 & 29 & - & 30 \\
$19\times 19$ & 38 & - & - & 38 \\
\bottomrule
\end{tabular}
\end{table}

\textbf{Integer Linear Programming.} ILP using Gurobi achieved provably optimal solutions across all tested grid sizes, successfully scaling to $n = 19$ where it placed 38 points. The branch-and-bound algorithm with constraint pruning enables exact optimization for grids where the $O(n^3)$ constraint space remains tractable. ILP provides the strongest baseline, demonstrating that traditional optimization methods remain highly effective for well-structured combinatorial problems. Figure \ref{fig:ilp_n19} shows the optimal $19 \times 19$ configuration, illustrating the complex geometric structure required for maximum point placement while satisfying all collinearity constraints.

\textbf{PatternBoost.} The transformer-based approach achieved optimal solutions on grids up to $n = 14$ (28 points), matching known optimal values despite never observing optimal training examples. On $n = 15$ grids, PatternBoost discovered 29-point configurations, demonstrating modest generalization beyond its training distribution. The method successfully learned to recombine geometric patterns from greedy-generated examples, achieving competitive performance without exhaustive search. However, PatternBoost did not scale beyond $n = 15$, suggesting limitations in pattern generalization for larger grid sizes. The training dynamics presented in Section 3.3 reveal how iterative refinement and pattern learning enable this performance through progressive improvement across 20 generations.

\textbf{Proximal Policy Optimization.} PPO achieved perfect solutions on $n = 10$ grids (20 points with zero violations), successfully learning valid placement policies through reward-based feedback. On $n = 11$ grids, the agent placed 22 points but incurred a single collinearity violation, indicating difficulty maintaining long-term constraint satisfaction as problem complexity increases. PPO demonstrated the ability to discover effective strategies from minimal problem structure (only the reward signal), but struggled to scale beyond small grid sizes where sequential planning becomes critical. Figures \ref{fig:rl_success} and \ref{fig:rl_failure} illustrate this sharp performance transition between successful constraint satisfaction at $n = 10$ and systematic failure at $n = 11$, with detailed analysis of failure modes provided in Section 3.4.

\subsection{Training Dynamics}

\begin{figure}[htbp]
	\centering
	\includegraphics[width=0.7\textwidth]{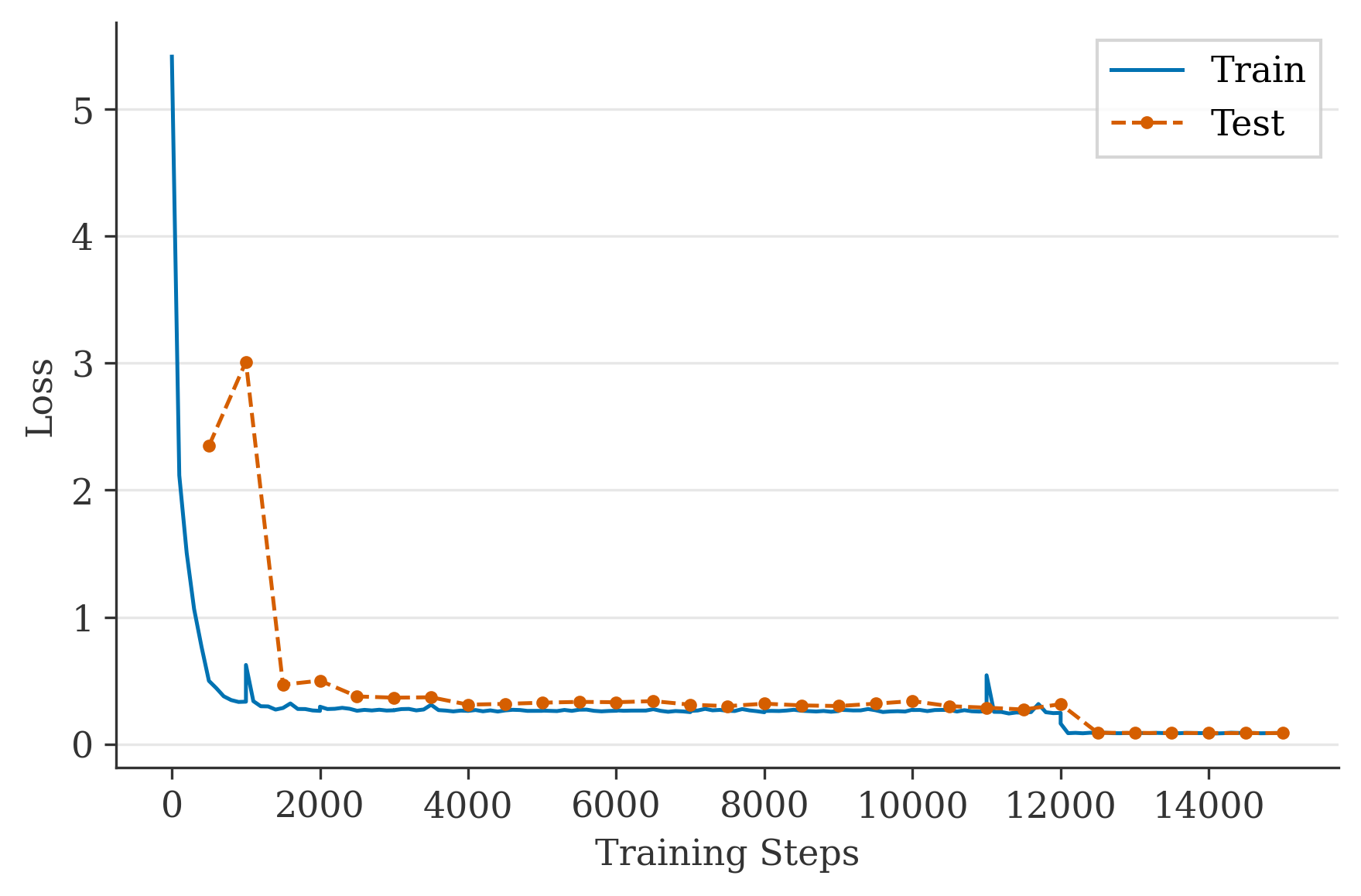}
	\caption{PatternBoost training and test loss over 15,000 steps.}
	\label{fig:patternboost_loss}
\end{figure}

Figure \ref{fig:patternboost_loss} shows PatternBoost's loss over 15,000 training steps. Training loss rapidly decreases from 5.5 to below 0.5 within the first 2,000 steps, demonstrating effective pattern learning from greedy-generated configurations. Test loss follows a similar trajectory, initially dropping from 3.0 to 0.5 by step 2,000, achieving a 96\% reduction over the training period. The minimal train-test gap throughout training indicates strong generalization without overfitting, confirming that the transformer successfully learns reusable geometric patterns rather than memorizing training examples. The small spike in test loss around step 12,000 corresponds to a generation transition where the training pool was updated with higher-quality configurations, temporarily increasing task difficulty before the model adapted. By the final generation, both train and test losses stabilize near 0.1, indicating convergence to patterns that reliably predict valid point placements.

\subsection{Limitations}

\begin{figure}[htbp]
	\centering
	\begin{minipage}{0.48\textwidth}
		\centering
		\includegraphics[width=\textwidth]{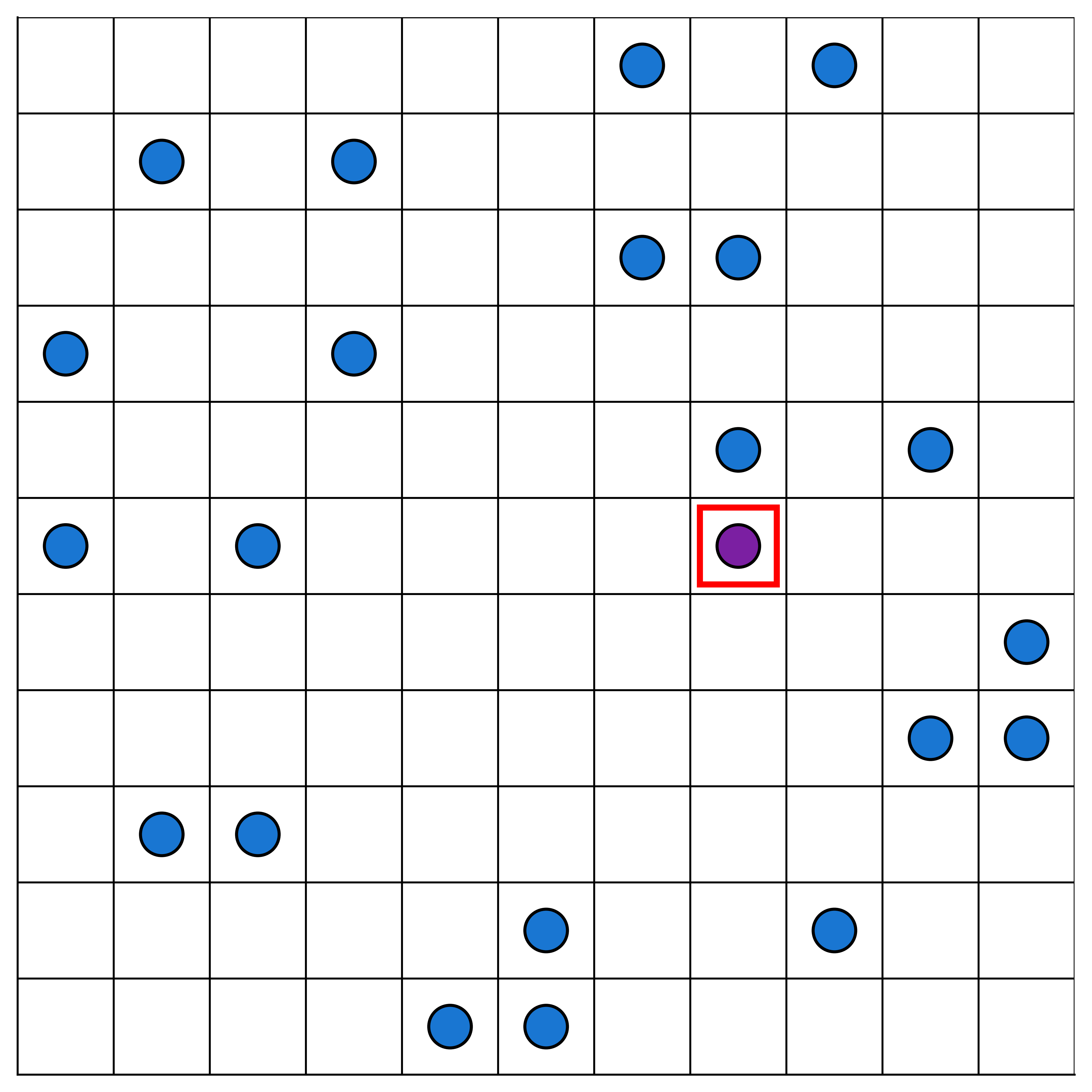}
		\caption{PPO failure on $11 \times 11$ grid: 22 points with 1 violation.}
		\label{fig:rl_failure}
	\end{minipage}
	\hfill
	\begin{minipage}{0.48\textwidth}
		\centering
		\includegraphics[width=\textwidth]{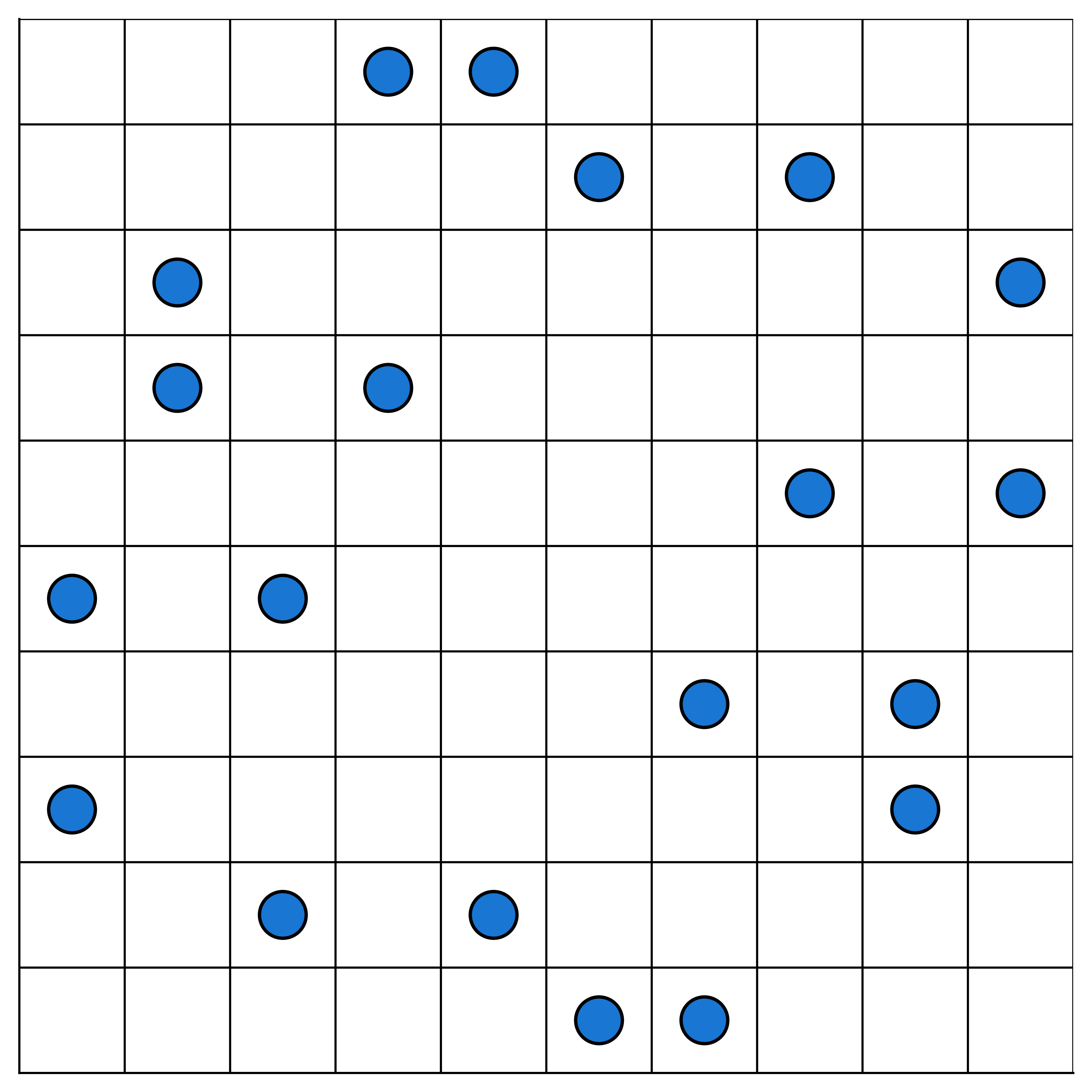}
		\caption{PPO success on $10 \times 10$ grid: 20 points with 0 violations.}
		\label{fig:rl_success}
	\end{minipage}
\end{figure}

Figures \ref{fig:rl_success} and \ref{fig:rl_failure} illustrate PPO's performance transition between successful and failed constraint satisfaction. On $10 \times 10$ grids (Figure \ref{fig:rl_success}), the agent consistently achieves perfect solutions with 20 points and zero violations, demonstrating successful policy learning for problems of this scale. However, on $11 \times 11$ grids (Figure \ref{fig:rl_failure}), despite placing 22 points, the agent incurs a single collinearity violation (highlighted point in red box), revealing the brittleness of learned policies as problem complexity increases. The violation occurs in a densely populated region where multiple collinearity constraints intersect, suggesting that the policy network struggles to maintain global constraint awareness when the number of placed points exceeds approximately 20. This represents a sharp transition from reliable performance to systematic failure as grid size increases by just one row and column.

\begin{figure}[htbp]
	\centering
	\includegraphics[width=0.6\textwidth]{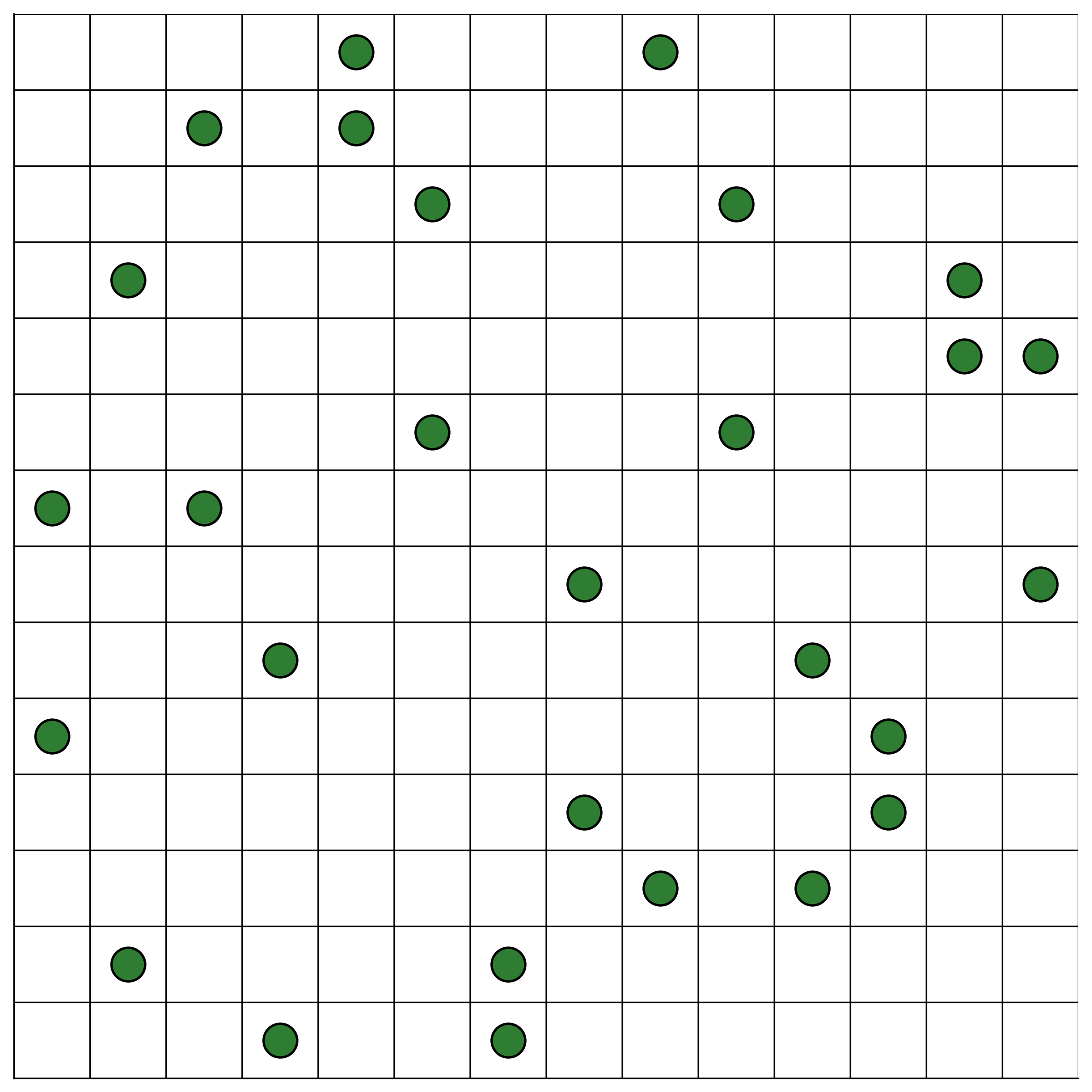}
	\caption{PatternBoost optimal solution on $14 \times 14$ grid achieving 28 points, demonstrating successful pattern learning from greedy-generated training data.}
	\label{fig:patternboost_n14}
\end{figure}

\begin{figure}[htbp]
	\centering
	\includegraphics[width=0.6\textwidth]{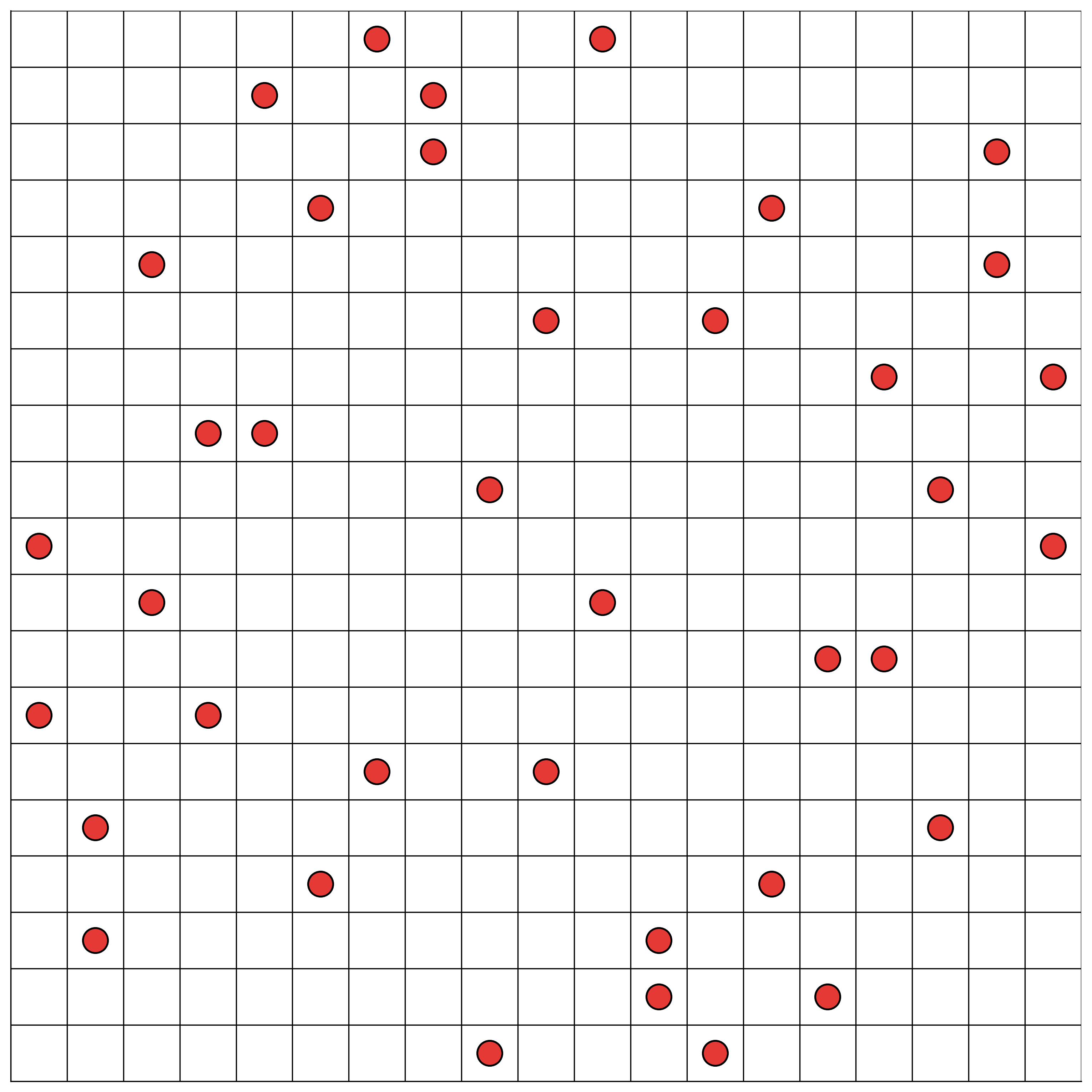}
	\caption{ILP optimal solution on $19 \times 19$ grid achieving 38 points.}
	\label{fig:ilp_n19}
\end{figure}

PatternBoost's failure at $n = 15$ follows a different pattern. While achieving 29 points on $15 \times 15$ grids, one point short of the known optimum of 30, the method demonstrates graceful degradation rather than catastrophic failure. The configurations discovered remain valid (zero violations) but fail to reach optimality, suggesting that the transformer's learned patterns generalize approximately but not perfectly to larger grids. Figure \ref{fig:patternboost_n14} shows PatternBoost's optimal $14 \times 14$ solution achieving 28 points, demonstrating the geometric patterns successfully learned through iterative refinement. Figure \ref{fig:ilp_n19} shows ILP's optimal $19 \times 19$ solution for comparison, illustrating the increased complexity that AI methods must learn to replicate at larger scales.

\section{Discussion and Conclusion}

Our study presents the first comparison of classical optimization and AI approaches to the No-Three-In-Line problem. Integer linear programming performed best, finding optimal solutions for up to $n = 19$, demonstrating the power of classical solvers for constraint satisfaction problems. In contrast, reinforcement learning approaches achieved perfect solutions only up to $n = 10$, with performance degrading by 42\% at $n = 11$ due to constraint violations. PatternBoost, the hybrid method combining classical greedy algorithms with transformers, achieved optimal solutions up to $n = 14$ and modest generalization to $n = 15$, bridging the gap between pure reinforcement learning and classical optimization. These findings underscore that hybrid approaches combining algorithmic heuristics with learned pattern recognition offer the most promising balance between solution quality and computational efficiency.

The distinct failure modes observed across methods reveal fundamental trade-offs in combinatorial optimization. ILP's branch-and-bound approach guarantees optimality through systematic search space pruning but faces exponential growth in problem complexity—constraints increased by 23\% from $n = 19$ (5,524 constraints) to $n = 20$ (6,778 constraints), illustrating the combinatorial explosion inherent to exact optimization. PatternBoost's philosophy of rapid training data generation through greedy algorithms creates a quality-speed trade-off: while fast data collection is central to the method's efficiency, the greedy approach struggles to find high-quality placements as problem complexity increases, constraining the quality of learned patterns. Reinforcement learning improved over training iterations but consistently produced constraint violations beyond $n = 10$, revealing fundamental difficulty in satisfying geometric constraints through reward-based exploration alone. These computational bottlenecks underscore that each method occupies a distinct niche in the optimization landscape.

This study is limited to grid sizes up to $n = 19$ for ILP, $n = 15$ for PatternBoost, and $n = 11$ for PPO, preventing definitive conclusions about asymptotic scaling behavior. We did not explore hybrid training combining supervised and reinforcement learning, transfer learning across grid sizes, or curriculum learning strategies that could leverage synergies between approaches. Computational constraints prevented exhaustive hyperparameter tuning and investigation of architectural improvements such as multi-branch networks or enhanced attention mechanisms for capturing long-range geometric dependencies. The PatternBoost model relies heavily on greedy-generated training data quality, with greedy algorithm limitations potentially propagating to the learned model. PPO's reward structure remains relatively simple and may benefit from more sophisticated shaping, intrinsic motivation signals, or exploration bonuses to better handle sparse binary rewards in constraint satisfaction tasks.

Future research should prioritize hybrid training paradigms that combine supervised pretraining (PatternBoost's cross-entropy learning) with reinforcement fine-tuning (PPO's exploratory optimization), mirroring successful approaches in language modeling. Transfer learning across grid sizes and curriculum strategies could enable training on smaller grids before scaling to larger instances, with meta-learning potentially discovering optimization strategies that generalize across problem scales. Classical-AI integration offers additional promising directions: ILP warm-starting with PatternBoost candidates could accelerate convergence on larger grids, while PatternBoost training on ILP-optimal smaller solutions could improve pattern quality beyond greedy baselines. Physics-inspired approaches present a particularly compelling direction, where combinatorial problems are reformulated as energy minimisation over relaxed Hamiltonians, enabling gradient-based neural network training~\cite{schuetz2022combinatorial}. 

\section*{Code Availability}

The code for all three methods is available in the following repositories:

\begin{itemize}
    \item ILP No-Three-In-Line: \url{https://github.com/pranav-ramanathan/N3L_gurobi}
    \item PatternBoost No-Three-In-Line: \url{https://github.com/pranav-ramanathan/N3L_PatternBoost}
    \item PPO No-Three-In-Line: \url{https://github.com/pranav-ramanathan/N3L_RL}
\end{itemize}

\section*{Acknowledgements}

This research was conducted as part STRIDE program at Queen Mary University of London. We thank the STRIDE team for their support, training, and development opportunities that made this work possible. We are particularly grateful for providing the resources and mentorship necessary to complete this project.

\bibliographystyle{unsrtnat}  
\bibliography{references}  


\end{document}